\documentclass[sigconf]{aamas} 

\usepackage{balance} 
\usepackage{hyperref}

\newenvironment{nohyphens}{%
  \par
  \hyphenpenalty=10000
  \exhyphenpenalty=10000
  \sloppy  
}{\par}

\acmConference[AAMAS '23]{Proc.\@ of the 22nd International Conference
on Autonomous Agents and Multiagent Systems (AAMAS 2023)}{May 29 -- June 2, 2023}
{London, United Kingdom}{A.~Ricci, W.~Yeoh, N.~Agmon, B.~An (eds.)}
\copyrightyear{2023}
\acmYear{2023}
\acmDOI{}
\acmPrice{}
\acmISBN{}

\acmSubmissionID{???}

\title[AAMAS-2023 Formatting Instructions]{%
  Synchronous vs Asynchronous Reinforcement Learning in a Real World Robot\\[0.5em]
  \large\textit{All authors contributed equally to this work.}%
}

\author{Ali Parsaee}
\affiliation{
  \institution{University of Alberta}
  \city{Edmonton, Alberta}
  \country{Canada}}
\email{parsaee@ualberta.ca}

\author{Chuxin He}
\affiliation{
  \institution{University of Alberta}
  \city{Edmonton, Alberta}
  \country{Canada}}
\email{chuxin@ualberta.ca}

\author{Fahim Shahriar}
\affiliation{
  \institution{University of Alberta}
  \city{Edmonton, Alberta}
  \country{Canada}}
\email{fshahri1@ualberta.ca}

\author{Ruiqing Tan}
\affiliation{
  \institution{University of Alberta}
  \city{Edmonton, Alberta}
  \country{Canada}}
\email{ruiqing2@ualberta.ca}

\usepackage{indentfirst}

\begin{abstract}
\begin{sloppypar}
\begin{nohyphens}
In recent times, reinforcement learning (RL) with physical robots has attracted the attention of a wide range of researchers. However, state-of-the-art RL algorithms do not consider that physical environments do not wait for the RL agent to make decisions or updates. RL agents learn by periodically conducting computationally expensive gradient updates. When decision-making and gradient update tasks are carried out sequentially by the RL agent in a physical robot, it significantly increases the agent's response time. In a rapidly changing environment, this increased response time may be detrimental to the performance of the learning agent. Asynchronous RL methods, which separate the computation of decision-making and gradient updates, are a potential solution to this problem. However, only a few comparisons between asynchronous and synchronous RL have been made with physical robots. For this reason, the exact performance benefits of using asynchronous RL methods over synchronous RL methods are still unclear. In this study, we provide a performance comparison between asynchronous and synchronous RL using a physical robotic arm called Franka Emika Panda. Our experiments show that the agents learn faster and attain significantly more returns using asynchronous RL. Our experiments also demonstrate that the learning agent with a faster response time performs better than the agent with a slower response time, even if the agent with a slower response time performs a higher number of gradient updates.

\end{nohyphens}
\end{sloppypar}
\end{abstract}

\setcopyright{none}
\renewcommand\footnotetextcopyrightpermission[1]{}

\begin{document}


\pagestyle{fancy}
\fancyhead{}


\maketitle 


\section{Introduction}

Robotic control is an important field with broad applications across multiple domains. Studies have shown the practical use of robotic control in medical, military, and space exploration tasks (Gyles, 2019; Russo \& Lax, 2022). Reinforcement learning (RL) is an area of machine learning often used for robotic control in online environments (Duan et al., 2016; Haarnoja et al., 2018; Schulman et al., 2015; Schulman et al., 2017; Yuan \& Mahmood, 2022). RL allows the robot to update its behaviors depending on its interaction with the environment, allowing it to adopt behaviors that better match the desired outcome. 

Most researchers use virtual environments instead of physical robots for robotic control using reinforcement learning (Yuan \& Mahmood, 2022). Traditional RL procedures are well-suited for virtual environments. However, there are challenges in using traditional RL procedures in real-time environments (Dulac-Arnold et al., 2020; Yuan \& Mahmood, 2022). One significant issue is that real-time environments do not pause when the robot makes decisions or updates, and traditional RL procedures do not take this into account. Inefficient or large computations by the robot may cause a discrepancy between the actions it takes and their respective outcomes, as well as a delayed reaction time for the robot relative to the environment (Yuan \& Mahmood, 2022). 


One way to improve a robot’s response time when using RL is asynchronous learning (Haarnoja et al., 2019; Yuan \& Mahmood, 2022). Asynchronous learning uses one process to sample actions and separate processes to collect batch data and make gradient updates. This setup considerably increases the response time of the robot, as sampling action is computationally very cheap compared to collecting batch data and making gradient updates. Yuan and Mahmood (2022) demonstrate the performance benefit of using asynchronous RL in a real-time robot. In our experiments, we hope to replicate the results of their work in a different physical robot, the Franka Emika Panda robotic arm. We hope to illustrate the effectiveness of asynchronous solutions in a new environment with similar challenges. 


We use the \textit{Franka-env}, an adaption of the code from Karimi et al. (2022), to handle the interactions between the robotic arm and the RL agent. We use the \textit{ReLoD} architecture  (Wang et al., 2022), which supports synchronous and asynchronous RL implementations using the soft actor-critic algorithm. To train the RL agents, we used images of different resolutions and the state information of the robotic arm. We conducted a large number of experiments to have a clear understanding of the performance of the RL agents with different settings.  

Our experiments show that the asynchronous RL variants accumulate higher average returns compared to the synchronous RL variants. We found that the primary reason for the asynchronous variants' better performance is the lower interaction time with the environment. Lower interaction time results in finer arm control, and the agents can also collect more samples per run. Before conducting the experiments, we thought that the asynchronous variants would perform better due to a higher number of gradient updates. However, the experiment results showed that this was not the case. Overall, we conducted 100 individual runs for the experiments, each taking 2 hours of wall clock time.

\section{RELATED WORKS}
\subsection{Reinforcement Learning with physical Robots}
Researchers use Reinforcement Learning widely in robotic tasks,
such as in racing slot cars (Lange et al., 2012), grasping objects (Kalashnikov et al., 2018),  robotic manipulation (Andrychowicz et al., 2020; Gu et al., 2017), and locomotion (Haarnoja et al., 2019). Proposed methods are applied to solve these tasks in real-word, such as zero constraint violation (Dalal et al., 2018), learning a value function for the reset policy (Eysenbach et al., 2018), rewards without hand-designed perception systems (Zhu et al., 2020), policy trained exclusively in simulation (Peng et al., 2018), and using auxiliary tasks (Schwab et al., 2019). Instead of synchronous learning, most of these works utilize asynchronous learning or distributed learning, which causes the need for more study between synchronous and asynchronous learning. Yuan and Mahmood (2022) studied the advantages of asynchronous learning over synchronous learning in real-world reinforcement learning.

\subsection{Asynchronous Reinforcement Learning}
Asynchronous Reinforcement Learning enables running environment interactions and updating parameters in a multi-threading way (Nair et al., 2015). Asynchronous learning allows each subsystem to run individually without affecting others, it makes it easier when one or more subsystems need to pause or restart (Haarnoja et al., 2018b). Asynchronous learning systems have been utilized in some real-world robotic control tasks, such as asynchronous deep reinforcement learning using the Normalized Advantage Function algorithm on real robotic platforms (Gu et al., 2017), SARSA learning algorithm on an asynchronous continuous-time robot task (Travnik et al., 2018), a systematic comparison between asynchronous learning and sequential learning on a robotic arm (Yuan et al. 2022). 


\section{Background}
\subsection{Markov Decision Process}
The Morkov Decision Process (MDP) is a mathematical model with finite states and actions. The continuous control task is formulated as a six-element finite-horizon Markov decision process \textit{M : (S, A, p, r, $\gamma$, $d_0$)}. In our experiments, \textit{S} and \textit{A} are continuous and represent the set of states called state space and the set of actions called action spaces respectively. 
The transition probability \textit{$p = Pr(s_{t+1}|s_t,a_t)$} is the probability that action \textit{$a_t$} $\in A$ in state \textit{$s_t$} $\in S$ at time \textit{t} will lead to \textit{$s_{t+1}$} $\in S$. The reward is a scalar feedback signal \textit{$r: S \times A \rightarrow R$}, discount factor $\gamma \in [0, 1)$, and $d_0$ represents the initial state distribution. 

\subsection{Soft Actor-Critic}
Off-policy Soft Actor-Critic (Haarnoja et al., 2018) attempts to find a policy $\pi_\phi(a_t|s_t)$ that maximizes the expected long-term reward and long-term entropy. We trained soft Q-function to minimize the following objective:
\begin{eqnarray}
J_Q(\theta) = \mathrm{E}_D[\frac{1}{2}(Q_\theta(s_t,a_t)-(R_{t+1}+\gamma V_{\hat{\theta}} (s_{t+1})))^2]
\end{eqnarray}
\begin{sloppypar}
where \textit{D} is the distribution of previously sampled states and actions, in our experiments, \textit{D} represents a replay buffer. $Q_\theta(s_t,a_t)$ is the soft action-value function parameter by $\theta$. Additionally, the soft-value $V_{\hat{\theta}} (s_{t+1})$ is sampled from $\mathrm{E}_{a_{t+1}\sim\pi_{\theta}}[Q_{\hat{\theta}}(s_{t+1},a_{t+1}) - \alpha\log\pi_{\phi}(a_{t+1}|s_{t+1})$, where $\alpha$ is the temperature parameter and $\hat{\theta}$ is the target network parameter. The new policy objective becomes:
\begin{eqnarray}
J_Q(\theta) = \mathrm{E}_D[\mathrm{E}_{a_{t}\sim\pi_{\theta}}[\alpha\log\pi_{\phi}(a_{t}|s_{t})-Q_{\theta}(s_{t},a_{t})]]
\end{eqnarray}

\end{sloppypar}
%

\balance

\section{The Environment Setup and the Task}

\begin{figure}
  \centering
  \includegraphics[width=1\linewidth]{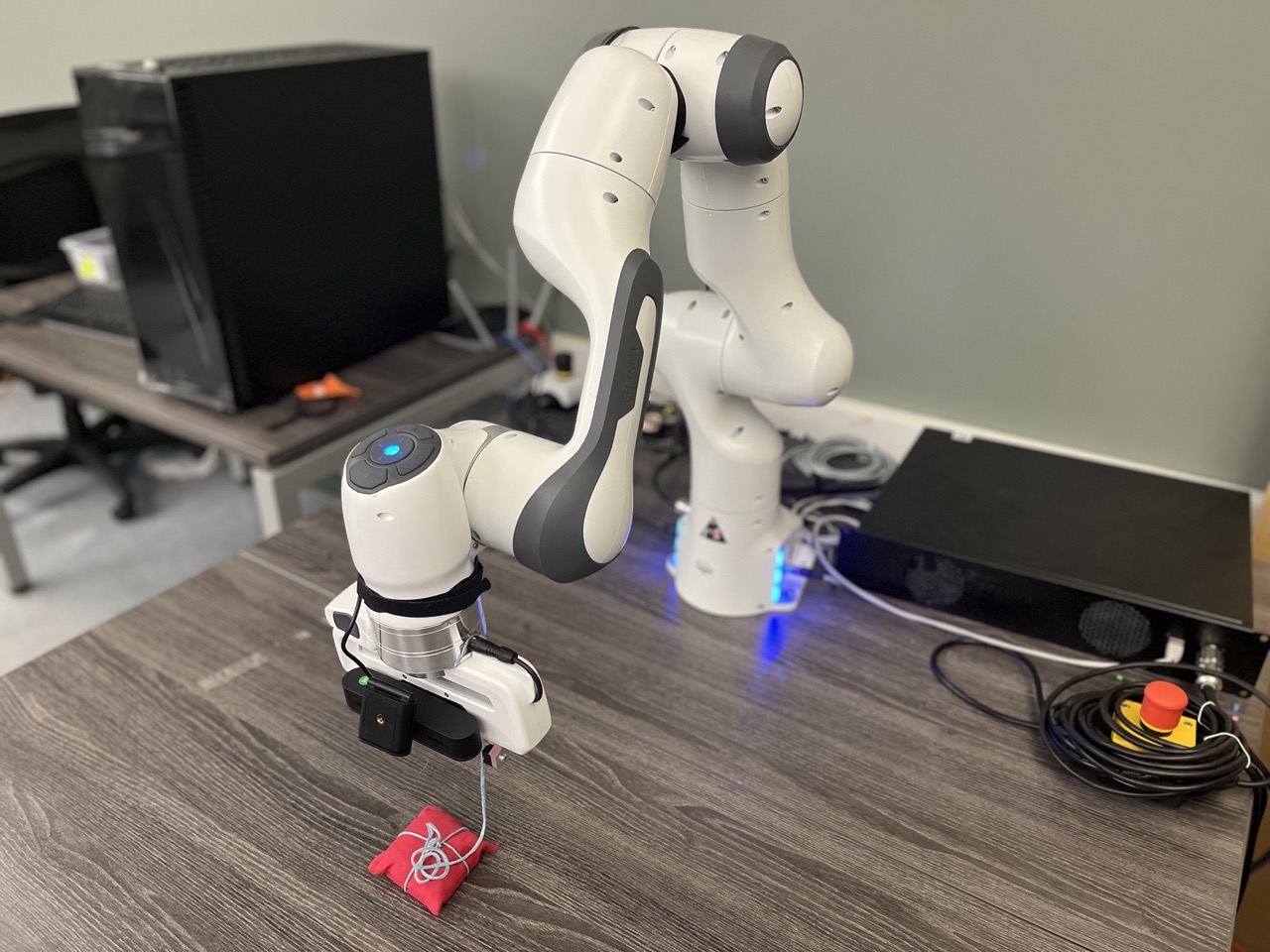}
  \caption{The Franka Emika Panda robotic arm}
  \label{fig:panda_arm}
\end{figure}

The learning environment consists of multiple hardware and software components. We have the Franka Emika Panda robotic arm, a Logitech webcam, and a small red bean bag for hardware. The Franka Emika Panda arm has seven joints with full torque sensing at each joint. We can control the arm by sending velocity commands for each joint. We do not use the end effector of the arm in our experiments. As shown in Figure \ref{fig:panda_arm}, we have attached the Logitech webcam at the tip of the arm. The bean bag is attached to the arm with a thread. This setting helps to place the bean bag in random positions on the table at the start of each episode. 

To handle the interaction between the learning agent and the hardware, we use an adaptation of the code from Karimi et al. (2022). We call this code the \textit{Franka-Env}. \textit{Franka-Env} is responsible for providing the current status of the arm and webcam images to the learning agent at each step. The current status of the arm consists of the current positions of the joints, current joint velocities, and the last velocity commands. Based on the setting, the size of an image is either 160×90×3 or 320×180×3. \textit{Franka-Env} provides the learning agent with a stack of three consecutive images for a better sense of direction. \textit{Franka-Env} receives velocity commands from the learning agent, a size seven vector, to control the seven joints of the robotic arm. The interaction between the learning agent and the \textit{Franka-Env} continually occurs after a certain period of time, which we call the action cycle time. We use different action cycle times for different settings in our experiments. To ensure the safety of the robotic arm, \textit{Franka-Env} bounds the movement of the arm to a box of dimensions 40×60×30 cm. We discuss the learning agent in section 5. 

\begin{figure}
  \centering
  \includegraphics[width=1\linewidth]{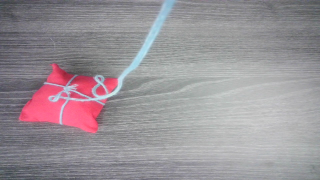}
  \caption{The webcam image, taken after a reset in the high-resolution setting}
  \label{fig:webcam}
\end{figure}

The task for our experiments is a vision-based control task, which we call the \textit{visual-reacher} task. The objective of the learning agent is to use the webcam images to control the arm to reach the red bean bag and remain close to the bag until the episode ends. We define the reward function as:

\begin{eqnarray}
R = \alpha \cdot \delta t \cdot \frac{1}{hw} \sum_{i=0}^{h}\sum_{j=0}^{w}M_{ij}   
\end{eqnarray}

Here, \(h\) and \(w\) are the height and width of the image. \(M\) is a 0-1 mask matrix with the same shape as the image. We set the pixel values to 1 in the mask matrix, in the positions where the bean bag is detected in the image. The rest of the pixel values are 0 in the mask matrix. We apply a color threshold filter on the image to create the mask matrix. We set \(\alpha\) to \(0.25\) and \(\delta t\) is the value of the action cycle time in milliseconds. The closer the webcam is to the bean bag, the larger the bean bag appears in the image, and the agent receives a higher reward. The agent can maximize the episodic return by moving close to the bean bag fast and remaining close to the bean bag until the episode ends. For all settings, we set the duration of an episode to 6 seconds and the number of episodes in an individual run to 720. After each episode, the arm resets to a fixed starting position in about 4 seconds. An individual run takes approximately 2 hours to complete. During this time, we use 72 minutes for training and the rest for resetting.

\section{Learning Architecture}
In this section, we describe the learning architecture, which mainly involves the implementation of the Synchronous Soft-Actor-Critic (SAC) and Asynchronous SAC. Similar to the Asynchronous RL framework published by Yuan and Mahmood (2022), our learning architecture consists of three processes: (1) the agent environment interaction process (AEIP), (2) the buffer sampling process, and (3) the gradient update process. In the agent environment interaction process, the agent makes action computations and samples an action from the current state: the joint angles, the joint velocity of the robotic arm, previous actions, and the stacked webcam images. The agent also collects the transition data, like the next state and reward, by applying the sampled action in the agent environment interaction process. Those transition data are stored in the buffer and sampled as mini-batches during the buffer sampling process. In the gradient update process, we update the gradient or policies. The synchronous SAC algorithm is strictly following the order of the above processes. 

\subsection{Asynchronous Implementation}
\begin{figure*}
  \centering
  \includegraphics[width=\linewidth]{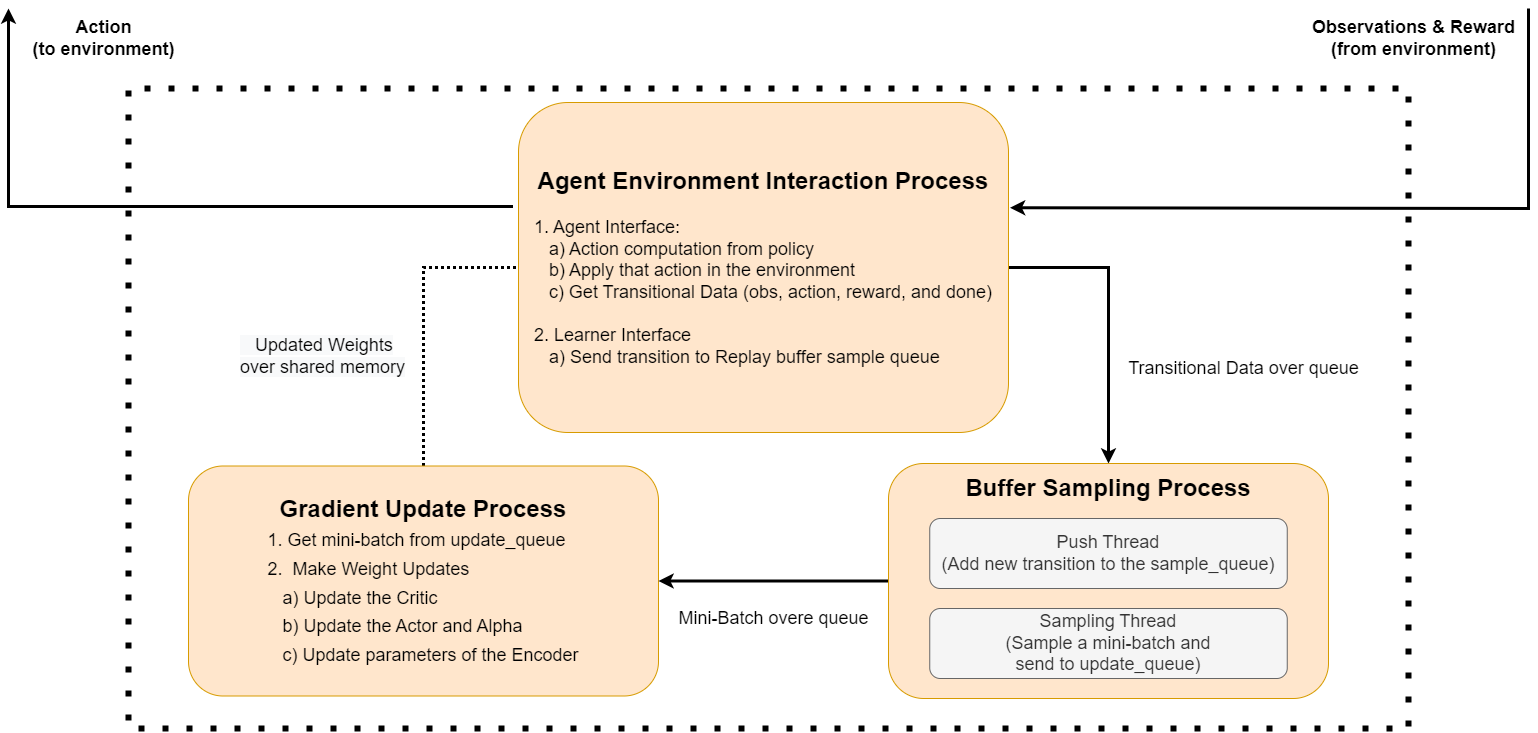}
  \caption{Overview of the Asynchronous Learning Architecture}
  \label{fig:Architect}
\end{figure*}

For the implementation of asynchronous SAC, we use the \textit{ReLoD} system. Wang et al. (2022) created the \textit{ReLoD} system to use RL with real-world robots. The \textit{ReLoD} system supports both synchronous and asynchronous learning, and we do not change the default hyper-parameter values of the \textit{ReLoD} system. We use the local mode of \textit{ReLoD}, which uses a local machine directly connected to the robot for learning. For our experiments, we are using a server equipped with an AMD Ryzen Threadripper 3970X 32-Core Processor and an NVIDIA GeForce RTX 3090. Under the local mode, the AEIP, the buffer sampling process, and the gradient update process all run on the local computer in an asynchronous and distributed manner. Figure \ref{fig:Architect} shows the asynchronous learning architecture. Transitional data collected in AEIP is pushed to the sample queue waiting for sampling, and the updated weights from the gradient update process are uploaded to AEIP over shared memory. This means the AEIP can keep interacting with the environment without waiting for the completion of buffer sampling and gradient update process. \textit{ReLoD} pushes the sampled mini-batches to an update queue to support the parallel running of buffer sampling and gradient updates.

\begin{figure}[htp!]
  \centering
  \includegraphics[width=1\linewidth]{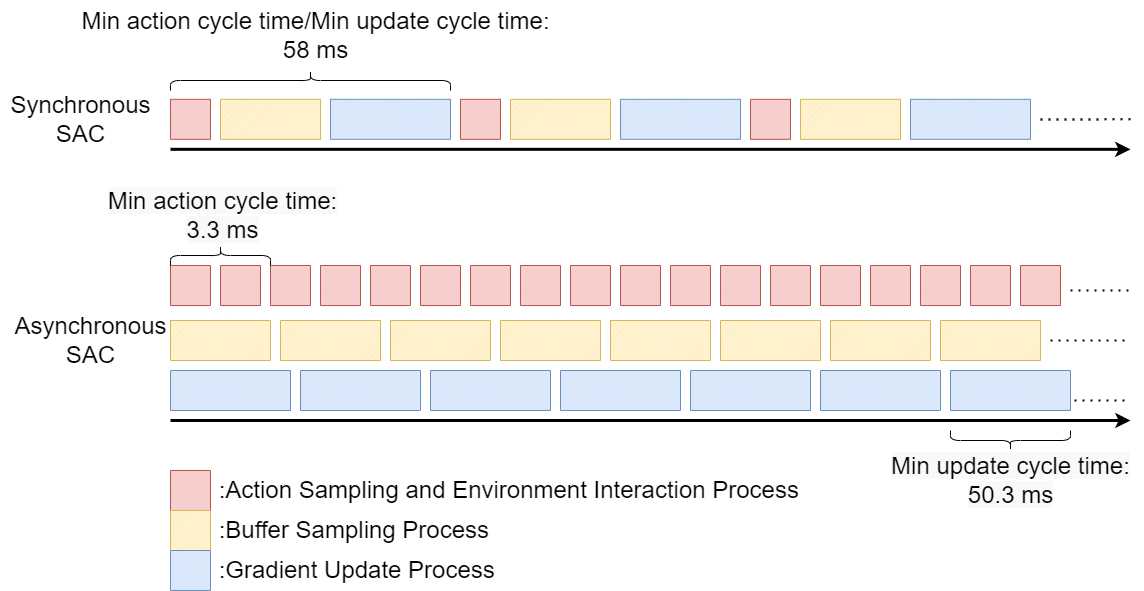}
  \caption{The Computational Flow of Synchronous SAC and Asynchronous SAC (The relative length of each block may not reflect the relative computation time)}
  \label{fig:ComFlow}
  
    \centering
  \includegraphics[width=1\linewidth]{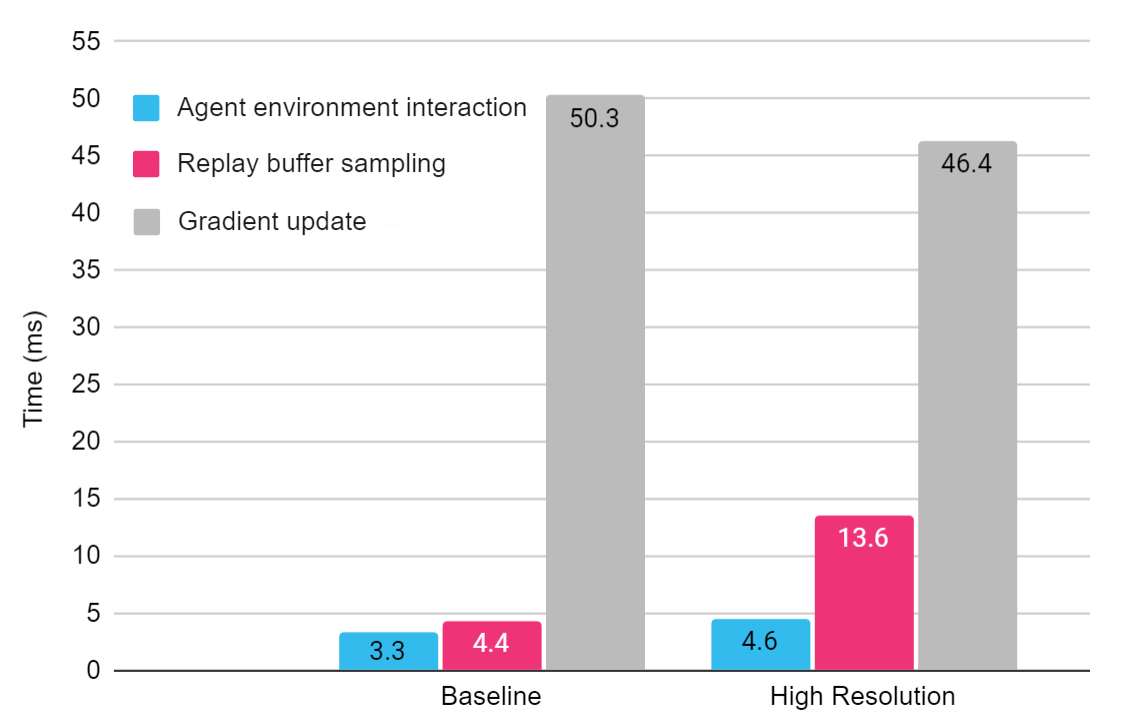}
  \caption{The computation time of different components in synchronous baseline and high-resolution settings}
  \label{fig:comput_time}
\end{figure}



To illustrate the benefits of asynchronous learning, we compare the computational flow of both synchronous and asynchronous variants in Figure \ref{fig:ComFlow} in the baseline setting. As we can see from the computational flows, for synchronous learning, the minimum action cycle time is equal to the minimum update cycle time, which is the sum of the time spent for each process. We record the computation time of individual components in the synchronous settings in Figure \ref{fig:comput_time}. In comparison, the action cycle time decreases significantly in asynchronous learning. This indicates that asynchronous architecture should perform better than synchronous one, especially when there is an expensive and huge amount of computation.



\section{The Experimental Setup}

We define four settings to compare the performance of synchronous and asynchronous learning. The first two settings are \textit{Asynchronous Baseline} and \textit{Synchronous Baseline} settings. For the baseline settings, we set the image size to 160×90×3 and the mini-batch size to 128. The following two settings are \textit{Asynchronous High-resolution} and \textit{Synchronous High-resolution} settings. We set the image size to 320×180×3 and the batch size to 80 for the high-resolution settings. For \textit{Asynchronous Baseline} and \textit{Asynchronous High-resolution} settings, we set the action cycle time to 40ms. For the \textit{Synchronous Baseline} setting, we set the action cycle time to 75ms, and for the \textit{Synchronous High-resolution} setting, we set the action cycle time to 80ms. Figure \ref{fig:comput_time} shows the average time required for different components during learning for the \textit{Synchronous Baseline} and \textit{Synchronous High-resolution} in a single step. The average values shown in Figure \ref{fig:comput_time} are computed using 164k samples for baseline and 154k samples for high-resolution settings. We opted for lower values of action cycle time for the synchronous settings, which do not hamper learning. When selecting the action cycle time, we left some room for \textit{Franka-Env} to prepare and provide the current status of the arm and webcam images. We found that for higher values of the action cycle time, at around 100ms, the arm frequently halts its movement. The halting of the arm hinders learning, and we consider this a limitation of the environment. Due to that reason, for the \textit{Synchronous High-resolution} setting, we used a smaller batch size of 80 to achieve a lower action cycle time of 80ms. 

We set the episode length to 6 seconds for all the settings. For both of the asynchronous settings, the number of steps in an episode is 150 due to the action cycle time of 40ms. For the  \textit{Synchronous Baseline} setting, the number of steps in an episode is 80 and for the \textit{Synchronous High-resolution} setting, the number of steps is 75. Due to the physical robot's constraints and the variable action cycle time, we made five modifications to the common experimental methodology. These modifications are based on the modifications proposed by Yuan and Mahmood, 2022. First, we set the training time to 72 minutes for all the tasks regardless of the number of steps the agent takes. Second, we scale the reward in proportion to the action cycle time (Eqn. 3). That is because, in a real-time setup, we have to specify the reward as a rate on a per unit time basis, such as ms (Doya, 2000). But because we cannot observe rewards continuously, we approximate the reward by setting it in proportion to the length of the action cycle time. Third, we provide the agent’s online performance for evaluation because the separate evaluation phase is incompatible with real-time learning setups. Fourth, we skipped hyper-parameter searching because the \textit{ReLoD} framework is already tuned, and hyper-parameter search might be harmful to the robot. Lastly, for the asynchronous variants, we used the initial 5000 steps out of 108,000 steps in each individual run for replay buffer initialization. We maintained the same ratio of initializing steps in synchronous variants. For \textit{Synchronous Baseline}, we used the first 2666 steps out of 57600 steps, and for \textit{Synchronous High-resolution}, we used the first 2500 steps out of 54000 steps for replay buffer initialization. 

\section{Results}
\begin{figure}[t]
  \centering
  \includegraphics[width=1\linewidth]{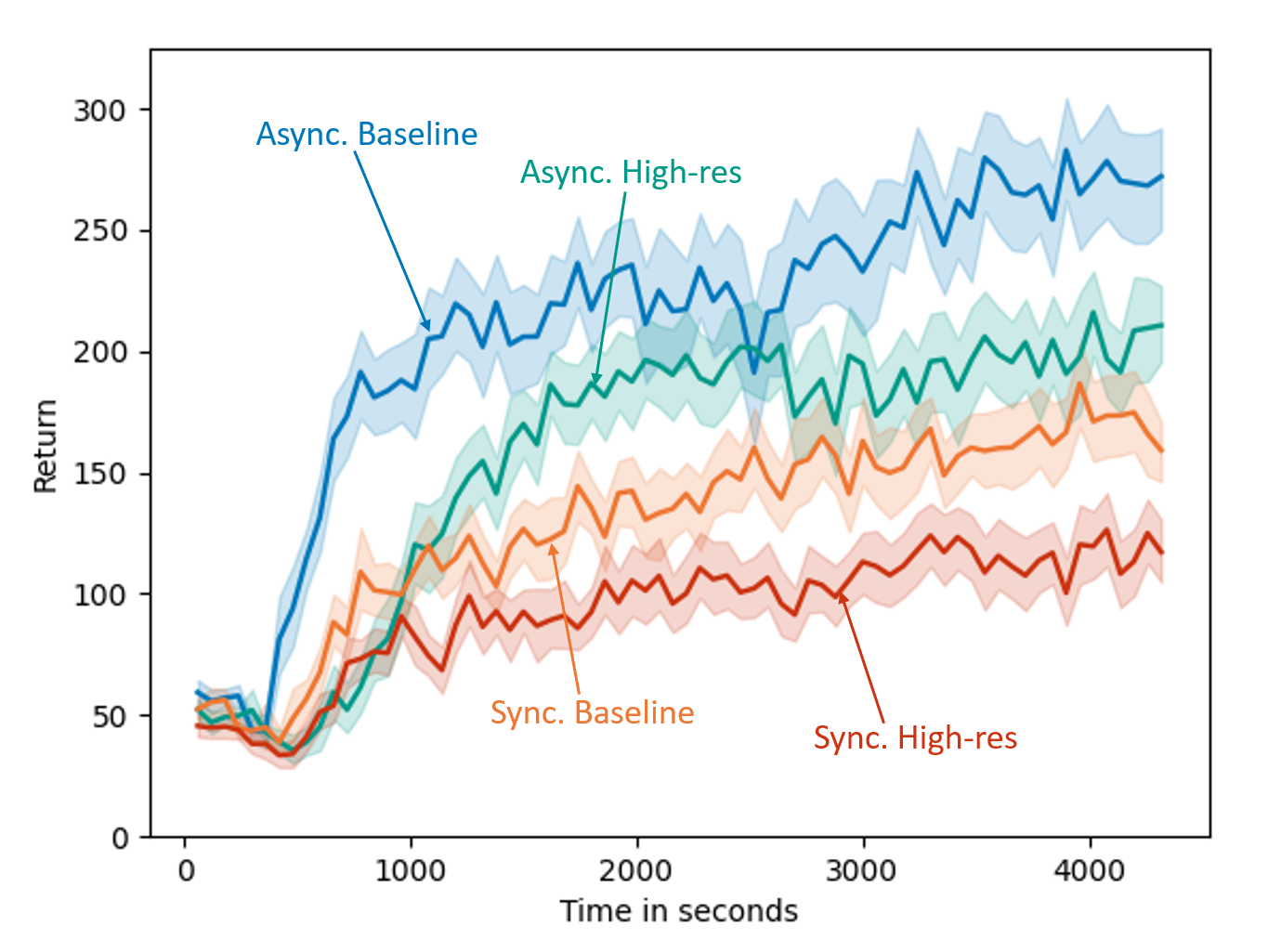}
  \caption{The learning curve of different settings averaged over twenty-five independent runs. The shaded regions correspond to 95\% confidence intervals of the averaged runs}
  \label{fig:learning_curve}

  \centering
  \includegraphics[width=1\linewidth]{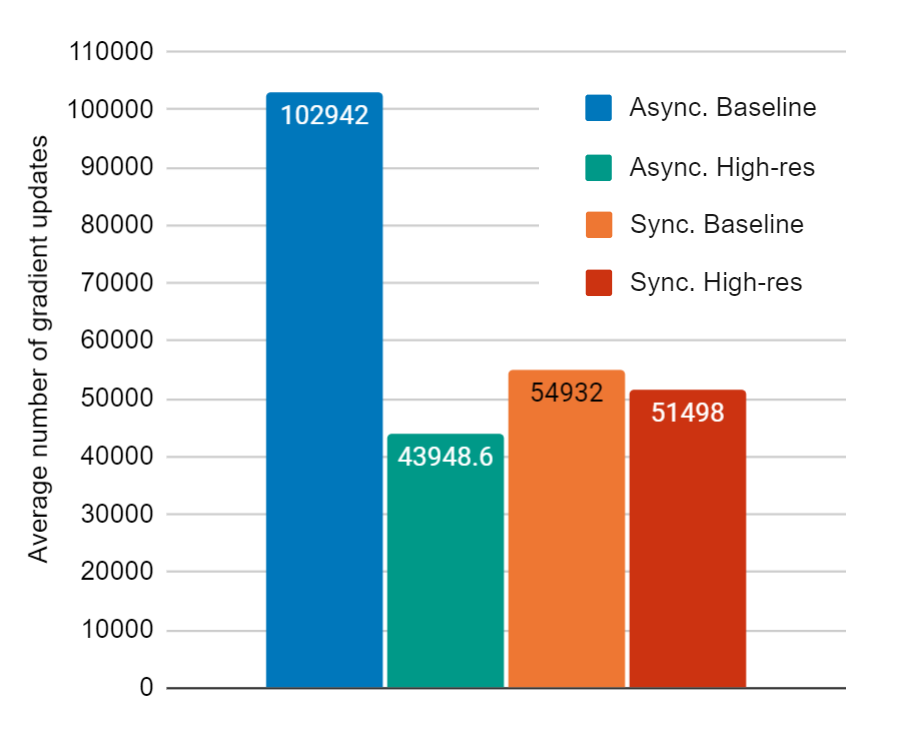}
  \caption{Average number of gradient updates per run for different settings}
  \label{fig:grad_update}
\end{figure}

We show the learning curves of \textit{Asynchronous Baseline}, \textit{Asynchronous High-Resolution}, \textit{Synchronous Baseline}, and \textit{Synchronous High-Resolution} settings in Figure \ref{fig:learning_curve}. We conducted twenty-five independent runs for each setting, resulting in 100 independent runs and 200 hours of learning with the Franka Emika Panda robotic arm. The experiments show that the performance of  \textit{Asynchronous Baseline} and \textit{Asynchronous High-Resolution} is significantly better than the synchronous counterparts. The \textit{Asynchronous High-Resolution} even managed to perform better than the \textit{Synchronous Baseline}. This fact becomes more interesting if we consider Figure \ref{fig:grad_update}. Figure \ref{fig:grad_update} shows that the \textit{Asynchronous High-Resolution} performed fewer learning updates than the synchronous variants. We conclude that the better performance of asynchronous versions is primarily due to lower action cycle time (or faster response time) and a higher number of samples collected for learning. The number of gradient updates has a less significant effect on performance. 

The \textit{Asynchronous High-Resolution conducting} a fewer number of gradient updates was not an expected behaviour. We reached out to the authors of \textit{ReLoD}, who also think this is unexpected behaviour. Finding the exact reason for this behaviour would require further experiments. However, we assume it happened due to inefficient data transfer between the buffer sampling and gradient update process. The synchronous settings, which sample from the replay buffer and perform gradient updates in the same process, do not face this phenomenon. 








\section{Conclusion}
In this work, we studied synchronous and asynchronous RL in a physical robot, the Franka Emika Panda robotic arm. We defined our task as the visual-reacher task and defined four settings for our experiments. We conducted 25 individual runs for each setting, a total of 100 individual runs. We showed that both \textit{Asynchronous Baseline} and \textit{Asynchronous High-Resolution} perform significantly better than the corresponding synchronous variants. We believe the asynchronous variants performed better due to a lower action cycle time (or faster response time) and a higher number of samples. Our experiments further showed that, even though \textit{Synchronous Baseline}, and \textit{Synchronous High-Resolution} had a higher number of gradient updates compared to the \textit{Asynchronous High-Resolution}, they did not perform better.

We want to extend our work in three distinct ways. First, we want to find out the reason behind fewer gradient updates in \textit{Asynchronous High-Resolution} and produce possible solutions to the problem. Second, we showed that lower action cycle time results in higher performance in the visual-reacher task with the robotic arm. We want to determine the best action cycle time for the given task. Lastly, we want to perform similar experiments with a different robot to solidify our hypothesis. 

\begin{acks}
\begin{nohyphens}
We want to thank Professor Adam White for his valuable guidance throughout the project. We also want to thank Professor Rupam Mahmood for suggesting the project and the Reinforcement Learning and Artificial Intelligence (RLAI) laboratory for providing the necessary hardware for conducting the experiments. Lastly, we want to thank Jordan Coblin and Vincent Liu for their encouraging comments and suggestions on the initial draft paper. 
\end{nohyphens}

\end{acks}

\balance
\section*{References}

\begin{sloppypar}
\hangindent=1em
\hangafter=1
\noindent Andrychowicz, M., Baker, B., Chociej, M.,  Józefowicz, R., McGrew, B., Pachocki, J., Petron, A., Powell, G., Ray, A., Schneider, J., Sidor, S., Tobin, J., Welinder, P., Weng, L., \& Zaremba, W. (2020). Learning dexterous in-hand manipulation. \textit{The International Journal of Robotics Research, 39}(1), 3–20. \href{https://journals.sagepub.com/doi/full/10.1177/0278364919887447}{https://journals.sagepub.com/doi/full/10.1177/0278364919887447}
\end{sloppypar}

\hangindent=1em
\hangafter=1
\noindent Dalal, G., Dvijotham, K., Vecerik, M., Hester, T., Paduraru, C., \& Tassa, Y. (2018). Safe exploration in continuous action spaces. \textit{arXiv preprint arXiv:1801.08757}. \href{https://arxiv.org/abs/1801.08757}{https://arxiv.org/abs/1801.08757}

\begin{sloppypar}
\hangindent=1em
\hangafter=1
\noindent Doya, K. (2000). Reinforcement learning in continuous time and space. \textit{Neural Computation, 12
(1):219–245}. \href{https://direct.mit.edu/neco/article-abstract/12/1/219/6324/Reinforcement-Learning-in-Continuous-Time-and?redirectedFrom=fulltext}{https://direct.mit.edu/neco/article-abstract/12/1/219/6324/Reinforcement-Learning-in-Continuous-Time-and?redirectedFrom=fulltext}
\end{sloppypar}

\hangindent=1em
\hangafter=1
\noindent Duan, Y., Chen, X., Houthooft, R., Schulman, J., \& Abbeel, P. (2016). Benchmarking deep reinforcement learning for continuous control. \textit{arXiv preprint arXiv:1604.06778}.\\
\href{https://arxiv.org/abs/1604.06778}{https://arxiv.org/abs/1604.06778}

\hangindent=1em
\hangafter=1
\noindent Eysenbach, B., Gu, S., Ibarz, J., \& Levine, S. (2018). Leave no trace: Learning to reset for safe and autonomous reinforcement learning. \textit{the International Conference on Learning Representations}. \href{https://openreview.net/forum?id=S1vuO-bCW}{https://openreview.net/forum?id=S1vuO-bCW}

\begin{sloppypar}
\hangindent=1em
\hangafter=1
\noindent Gu, S., Holly, E., Lillicrap, T., \& Levine, S. (2017). Deep reinforcement learning for robotic manipulation with asynchronous off-policy updates. \textit{2017 IEEE international conference on robotics and automation (ICRA)}, 3389–3396. \href{https://doi.org/10.1109/ICRA.2017.7989385}{https://doi.org/10.1109/ICRA.2017.7989385}
\end{sloppypar}

\begin{sloppypar}
\hangindent=1em
\hangafter=1
\noindent Gyles, C. (2019). Robots in medicine. \textit{The Canadian veterinary journal, v.60(8)}. \href{https://www.ncbi.nlm.nih.gov/pmc/articles/PMC6625162/}{https://www.ncbi.nlm.nih.gov/pmc/articles/PMC6625162/}
\end{sloppypar}

\hangindent=1em
\hangafter=1
\noindent Haarnoja, T., Zhou, A., Hartikainen, K., Tucker, G., Ha, S., Tan, J., Kumar, V., Zhu, H., Gupta, A., Abbeel, P., \& Levine, S. (2018). Soft actor-critic algorithms and applications. \\ \textit{arXiv preprint arXiv:1812.05905}. \href{https://arxiv.org/abs/1812.05905}{https://arxiv.org/abs/1812.05905}

\hangindent=1em
\hangafter=1
\noindent Haarnoja, T., Ha, S., Zhou, A., Tan, J., Tucker, G., \& Levine, S. (2018b). \textit{Learning to walk via deep reinforcement learning. arXiv preprint arXiv:1812.11103} \href{https://doi.org/10.48550/arXiv.1812.11103}{https://doi.org/10.48550/arXiv.1812.11103}

\hangindent=1em
\hangafter=1
\noindent Haarnoja, T., Ha, S., Zhou, A., Tan, J., Tucker, G., \& Levine, S. (2019). Learning to walk via deep reinforcement learning. \textit{Robotics: Science and Systems}.
\href{https://arxiv.org/abs/1812.11103}{https://arxiv.org/abs/1812.11103}

\hangindent=1em
\hangafter=1
\noindent  Karimi, A., \& Mahmood, A. R. Franka Visual Reacher, \textit{GitHub}, \href{https://github.com/amir-karimi96/franka\_visual\_reacher.git}{https://github.com/amir-karimi96/franka\_visual\_reacher.git}

\hangindent=1em
\hangafter=1
\noindent Kalashnikov, D., Irpan, A., Pastor, P., Ibarz, J., Herzog, A.,
Jang, E., Quillen, D., Holly, E., Kalakrishnan, M., Vanhoucke, V., \& Levine, S. (2018). Qt-opt: Scalable deep reinforcement learning for vision-based robotic manipulation. \textit{the Conference on Robot Learning}. \href{https://arxiv.org/abs/1806.10293}{https://arxiv.org/abs/1806.10293}

\hangindent=1em
\hangafter=1
\noindent Lange, S., Riedmiller, M., \& Voigtlander, A. (2012). Autonomous reinforcement learning on raw visual input data in a real world application. \textit{the International Joint Conference on Neural Networks}. \href{https://ieeexplore.ieee.org/document/6252823}{https://ieeexplore.ieee.org/document/6252823}

\hangindent=1em
\hangafter=1
\noindent Nair, A., Srinivasan, P., Blackwell, S., Alcicek, C., Fearon, R., De Maria, A.,
Panneershelvam, V., Suleyman, M., Beattie, C., Petersen, S., et al. (2015).
Massively parallel methods for deep reinforcement learning. \textit{arXiv preprint arXiv:1507.04296.}
\\
\href{https://doi.org/10.48550/arXiv.1507.04296}{https://doi.org/10.48550/arXiv.1507.04296}

\hangindent=1em
\hangafter=1
\noindent Peng, X. B., Andrychowicz, M., Zaremba, W., \& Abbeel, P. (2018). Sim-to-real transfer of robotic control with dynamics randomization. \textit{2018 IEEE international conference on robotics and automation (ICRA)}. \href{https://ieeexplore.ieee.org/document/8460528}{https://ieeexplore.ieee.org/document/8460528}

\hangindent=1em
\hangafter=1
\noindent Russo, A., \& Lax, G. (2022). Using artificial intelligence for space challenges: A survey. \textit{ Multidisciplinary Digital Publishing Institute(MDPI) open access journals, 12(10)}. \href{https://www.mdpi.com/2076-3417/12/10/5106/htm}{https://www.mdpi.com/2076-3417/12/10/5106/htm}

\hangindent=1em
\hangafter=1
\noindent Schulman, J., Levine, S., Abbeel, P., Jordan, M., \& Moritz, P. (2015). Trust region policy optimization.\textit{International conference on machine learning (pp. 1889–1897)}. \href{https://arxiv.org/abs/1502.05477}{https://arxiv.org/abs/1502.05477}

\begin{sloppypar}
\hangindent=1em
\hangafter=1
\noindent Schulman, J., Wolski, F., Dhariwal, P., Radford, A., \& Klimov, O. (2017). Proximal policy optimization algorithms. \mbox{\textit{arXiv preprint arXiv:1707.06347}}.\href{https://arxiv.org/abs/1707.06347}{https://arxiv.org/abs/1707.06347}
\end{sloppypar}

\hangindent=1em
\hangafter=1
\noindent Schwab, D., Springenberg, T., Martins, M. F., Lampe, T., Neunert, M., Abdolmaleki, A., Hertweck, T., Hafner, R., Nori, F., \& Riedmiller, M. (2019).
Simultaneously learning vision and feature-based control
policies for real-world ball-in-a-cup. \textit{Robotics: Science and Systems}. \href{https://arxiv.org/abs/1902.04706}{https://arxiv.org/abs/1902.04706}

\hangindent=1em
\hangafter=1
\noindent Travnik, J. B., Mathewson, K. W., Sutton, R. S., Pilarski, P. M. (2018). Reactive reinforcement learning in asynchronous environments. \mbox{\textit{Frontiers in Robotics and AI, 5.}}
\\\href{https://doi.org/10.3389/frobt.2018.00079}{https://doi.org/10.3389/frobt.2018.00079}

\hangindent=1em
\hangafter=1
\noindent Wang, Y., Vasan, G., \& Mahmood, A. R. (2022). Real-time reinforcement learning for vision-based robotics utilizing local and remote computers. \textit{arXiv preprint arXiv:2210.02317 }. \\\href{https://arxiv.org/abs/2210.02317}{https://arxiv.org/abs/2210.02317}

\begin{sloppypar}
\hangindent=1em
\hangafter=1
\noindent Yuan, Y., \& Mahmood, A. R. (2022). Asynchronous reinforcement learning for real-time control of Physical Robots. \textit{arXiv preprint arXiv:2203.12759}. \href{https://arxiv.org/abs/2203.12759}{https://arxiv.org/abs/2203.12759}
\end{sloppypar}

\hangindent=1em
\hangafter=1
\noindent Zhu, H., Yu, J., Gupta, A., Shah, D., Hartikainen, K., Singh,
A., Kumar, V., \& Levine, S. (2020). The ingredients of realworld robotic reinforcement learning. \textit{the International
Conference on Learning Representations}. \href{https://openreview.net/forum?id=rJe2syrtvS}{https://openreview.net/forum?id=rJe2syrtvS}

\hangindent=1em
\hangafter=1
\noindent Bertsekas, D.P (1983). Distributed asynchronous computation of fixed points. \textit{Mathematical Programming 27, 107–120 (1983)}. \\
\href{https://doi.org/10.1007/BF02591967}{https://doi.org/10.1007/BF02591967}


\end{document}